\documentclass{article} 
\usepackage{iclr2026_conference,times}
\usepackage[linesnumbered,ruled,vlined]{algorithm2e} 

\usepackage{amsmath,amsfonts,bm}

\def\eqref#1{equation~\ref{#1}}

\def\1{\bm{1}}

\DeclareMathAlphabet{\mathsfit}{\encodingdefault}{\sfdefault}{m}{sl}
\SetMathAlphabet{\mathsfit}{bold}{\encodingdefault}{\sfdefault}{bx}{n}

\usepackage{hyperref}
\usepackage{url}
\usepackage{amsmath}       
\usepackage{makecell}
\usepackage{bm}
\usepackage{graphicx}
\usepackage{enumitem}   
\usepackage{multicol}   
\usepackage{tabularx}
\usepackage{booktabs} 
\usepackage{marvosym}
\usepackage{footmisc}

\title{ToMPO: Training LLM Strategic Decision \\ Making from a Multi-Agent Perspective}

\author{Yiwen Zhang$^\dagger$$^{1,3}$, Ziang Chen$^\dagger$$^{1,4}$, Fanqi Kong$^{1,2}$, Yizhe Huang$^{1,2}$, Xue Feng $^{\text{\Letter}1}$ \\
$^1$State Key Laboratory of General Artificial Intelligence, BIGAI 
$^2$Peking University \\
$^3$The Hong Kong University of Science and Technology (Guangzhou) 
$^4$Tsinghua University
\\
\texttt{yzhang452@connect.hkust-gz.edu.cn, fengxue@bigai.ai}\\
}

\iclrfinalcopy 
\begin{document}

\maketitle

\footnotetext[0]{$^\dagger$Equal contribution. \text{\Letter} Corresponding author.}
\begin{abstract}

Large Language Models (LLMs) have been used to make decisions in complex scenarios, where they need models to think deeply, reason logically, and decide wisely.
Many existing studies focus solely on multi-round conversations in social tasks or simulated environments, neglecting the various types of decisions and their interdependence.
Current reinforcement learning methods struggle to consider the strategies of others during training.
To address these issues, we first define a strategic decision-making problem that includes two types of decisions and their temporal dependencies. 
Furthermore, we propose \textbf{T}heory \textbf{o}f \textbf{M}ind \textbf{P}olicy \textbf{O}ptimization \textbf{(ToMPO)} algorithm to optimize the perception of other individual strategies and the game situation trends.
Compared to the Group Relative Policy Optimization (GRPO) algorithm, ToMPO enhances the LLM's strategic decision-making mainly by: 1) generating rollouts based on reasoning the strategies of other individuals, 2) estimating advantages at both the graph-level and sample-level, and 3) balancing global and partial rewards.
The ToMPO algorithm outperforms the GRPO method by 35\% in terms of model output compliance and cooperative outcomes. Additionally, when compared to models with parameter sizes 100 times larger, it shows an 18\% improvement. This demonstrates the effectiveness of the ToMPO algorithm in enhancing the model's strategic decision-making capabilities.

\end{abstract}

\section{Introduction}


Large Language Models (LLMs) utilize natural language understanding and generation capabilities to achieve leading performance in decision-making scenarios, assisting people in generating \citep{gou2024rationality}, simulating \citep{mao2025alympics}, and predicting \citep{zhang2024electionsim} decisions across various categories.
While LLMs excel in coding and math tasks, they struggle with strategic decision-making, which requires understanding others' intentions, predicting behaviors, and adjusting their own strategies dynamically \citep{zhang2024llm}.

LLMs demonstrate varying strategic abilities in matrix games \citep{lore2024strategic,herr2024large} and can be enhanced through a game-theoretic workflow \citep{hua2024game}.
Recent research further explores LLM strategic decision-making through multi-level thinking \citep{zhang2024k,gou2024rationality}, Theory of Mind \citep{duan2024reta,crosshypothetical}, task-solving \citep{zhang2025sotopia,wang2024sotopia}, as well as influences between individuals and groups \citep{mi2025mf,zhang2025attention}. (detailed related work in section \ref{relatedworks})
These studies provide methods for LLMs to adapt to human society, emerge human behaviors, and serve social issues.
However, these studies restrict the strategic decision-making capabilities of LLM to two-agent chatroom environments or single-game scenarios. 
This approach fails to provide the necessary methods for LLM to enhance its performance in diverse, long-term multi-agent decision-making tasks.

By focusing on these key issues, our paper analyzes the strategic decision-making capabilities of LLMs in complex social environments, where LLMs must sequentially make decisions that impact both individuals and groups.
During this period, the prior decisions made by LLM will have a certain degree of influence on subsequent decisions. 
This implies that individual behaviors may lead to changes in the social structure of the group, and at the same time, changes in the group structure will affect subsequent individual decisions.
For instance, in real life, before signing a cooperation agreement with multiple distributors, enterprises will conduct various evaluations. 
After the cooperation agreement is signed, they will implement the cooperation with varying levels of investment over a specified period. 
Each cooperation has a certain impact on whether the enterprise decides to continue the next collaboration.
Furthermore, if there is a desire to terminate the cooperation during the process, it cannot be done immediately; that is, reversing the decision is not possible.
This decision-making process helps highlight the real-world challenges faced by individuals and groups over time, posing a challenge to the model's capabilities.

In this context, we first define the problem as a sequential decision-making process that primarily involves graph-level and effort-level decisions.
Then we build three kinds of complex social environments to test SOTA (State-of-the-Art) LLM performance. 
To optimize performance, we propose a reinforcement learning algorithm that integrates a multi-agent perspective into the LLM-based policy model training process.
Based on the preliminary tests, we created an expert dataset containing the effort-level decisions made by models that achieve high rewards, across various topological positions and at different stages in the game.
The policy model effectively learns decision-making at the effort level from the expert dataset through a supervised fine-tuning process. 
We enhanced the policy model for graph-level decision-making through reinforcement fine-tuning, which incorporates multi-agent considerations in reward modeling during the training process. 

Our contribution can be summarized as:

\begin{itemize}
    \item We define a problem for real-world strategic decision-making and design corresponding general simulation environments for decision data generation and examination.
    
    \item We evaluate the performance of the State-of-the-Art (SOTA) models and provide a dataset including the expert model's strategic decisions under different topological structures and at different game time processes.

    \item We propose a reinforcement learning algorithm, Theory of Mind Policy Optimization (ToMPO), and apply it to the \texttt{Qwen-2.5-7B-instruct} model, achieving improvement in strategic decision-making capabilities.
\end{itemize}

\section{Problem Formulation}


In contrast to the scenarios discussed in Theory of Mind \citep{strachan2024testing,liu2025mind} and single LLM long-term planning \citep{huang2024understanding,manon}, we require the LLM to operate as an agent within a multi-agent environment consisting of at least three agents, making two types of decisions sequentially.
During any decision-making process, an agent considers the strategies of other agents and its subsequent strategy, depending on its own state. 
These considerations will autonomously change based on the agent's social status, game progress, and others' performance.

 \paragraph{Graph-Effort Strategic Decision-Making} 


We define the decision-making process as a set $\langle \mathcal{N}, \mathcal{S},\mathcal{A}, T, \tau, f, r ,\gamma \rangle$, with the set of all agents or players $\mathcal{N} = \{ 1, 2, \dots, N \}$, state space $\mathcal{S}$, total game round $T$, decision type sequence $\tau$, the state transition function $f \in \{f_G, f_E\}$, utility function $r$, and discount factor $\gamma$.
The agent state at round $t$ includes the agent's social relationship structure ($\mathcal{G}$) and effort ($\mathcal{E}$) at this round. $L$ and $M$ represent how many steps of actions related to structure forming and effort investment, respectively in one round. $\tau$ represents the overall action type sequence. For example, when $L=1, M=2$, let \( \tau = \{(G, E, E), (G, E, E), \ldots\} \) represent a sequence where the LLM agent must make a graph-level decision at step 0 in one game round. This decision involves choosing whether to join one group or establish relationships with others. At steps 1 and 2, the agent will determine how much to invest based on the social relations established in step 0. This pattern continues in subsequent rounds. $a_{i,{t,j}}$ is the action decision agent $i$ made at step $j$ of round $t$  (equation \ref{a_t}). Action space $\mathcal{A} = \{A_i\}_{i \in \mathcal{N}}=  \{a_{i,\tau}\}_{i \in \mathcal{N}} $ (equation \ref{A_t}).
\begin{equation}
    \forall i \in \mathcal{N},\,\,  t \in [0, T-1],\,\,  j \in [0, L+M-1]
\end{equation}
\begin{equation}
    S_t = (G_t, E_t), \quad
    \tau(t,j) \in \{\text{G}, \text{E}\}, \quad
    a_{i,{t,j}} = 
    \begin{cases} 
        g_{i,{t,j}} & \text{if } \tau(t,j) = \text{G} \\
        e_{i,{t,j}} & \text{if } \tau(t,j) = \text{E}
    \end{cases}
    \label{a_t}
\end{equation}
\begin{equation}
    A_t = (A_G^{t,L}, A_E^{t,M}) = (\{ A_G^{t,0}, A_G^{t,1}, ..., A_G^{t,{L-1}}\}, \{ A_E^{t,{L}}, A_E^{t,{L+1}}, ..., A_E^{t,{L+M-1}} \})
    \label{A_t}
\end{equation}

\paragraph{Decision-Making Optimization with Credit Assignment} According to Credit Assignment \citep{nguyen2018credit,pignatellisurvey} in reinforcement learning, we decompose strategic decision-making into dual complementary processes (equation \ref{eq:forward} and \ref{eq:inverse}). 
$V^*$ represents the optimal value function, and $\mathcal{H}_t = \{ (a_\tau, r_\tau) \}_{\tau=0}^{t-1}$ denotes the decision-reward history. 
The forward process is designed to optimize the model's decision-making capabilities within a defined social graph structure. 
It effectively involves understanding the rules, accurately predicting or assessing the decisions of other agents, and clearly defining its own strategy.
On the other hand, the inverse process significantly enhances the model's ability to determine which group structure it will join next, relying on its memory of past decisions.
These two processes align with the credit assignment principle (equation \ref{creditass}).
\begin{align}
\text{Forward Process (Effort Decision Optimization):} & \quad \max_{e_t} \mathbb{E} \left[ \sum_{k=t}^T \gamma^{k-t} r_k \mid G_t = g \right] \label{eq:forward} \\
\text{Inverse Process (Graph Decision Optimization):} & \quad \max_{g_t} \mathbb{E} \left[ V^*(S_{t+1}) \mid \mathcal{H}_t \right] \label{eq:inverse}
\end{align}
\begin{equation}
\nabla_{\theta}J(\theta) = \underbrace{\sum_{t:\tau(t)=\mathcal{E}} \psi_E(\delta_t)}_{\text{\makecell{Forward Process Credit}}} + \underbrace{\sum_{t:\tau(t)=\mathcal{G}} \psi_G(\delta_t)}_{\text{\makecell{Inverse Process Credit}}} + \zeta(\Delta \mathcal{C})
\label{creditass}
\end{equation}



\section{Preliminary Testing LLM Strategic Decision-Making}

\subsection{Graph-Effort Strategic Game Design}
\begin{figure}[ht]
    \centering
    \includegraphics[width=\linewidth]{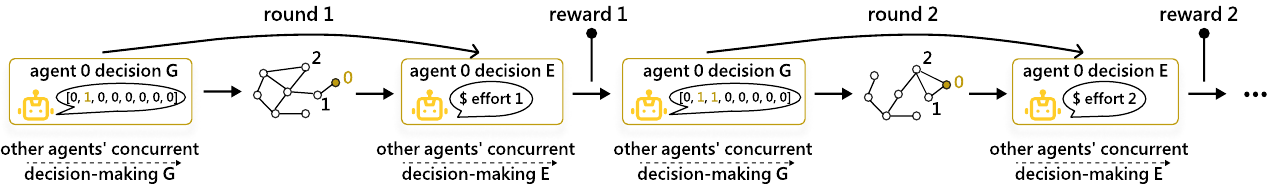}
    \caption{Demonstration of a two-round decision-making process in the GE sub-environment.}
    \label{fig:GEcase}
\end{figure}

We present two sequential multi-agent game environments where Large Language Model (LLM) agents make decisions over $T$ rounds. Both environments involve $N$ agents making choices related to social graph formation (G) and effort investment (E) to maximize their individual utility. Agents observe full historical information ($G_{history}, x_{history}, \pi_{history}$) to inform their current decisions. The decision-making process in each round generally consists of two key components: graph formation and effort investment. 
An agent refers to an individual who participates in the game and is part of the graph.
As shown in Figure \ref{fig:GEcase}, the agent, represented by the policy model, makes decisions simultaneously as other agents in the environment.

\subsubsection{Sequential \textit{BCZ} Game}
This environment extends the classic \textit{BCZ} (Bala-Goyal-Jackson) game \citep{ballester2006s} to a sequential framework. Each agent $i$ simultaneously decides on their social links and effort investments.
The sequence in which these decisions are made defines three sub-environments: GE, GEE, and GGE (detailed in Appendix \ref{appendix:env_bcz_sub}).

\textbf{Decision Components}

\begin{itemize}[leftmargin=0pt, itemindent=10pt, labelsep=5pt]
\item \textbf{Link Decision(G):} All agents simultaneously decide on mutual social links, represented by an adjacency matrix $G \in \{0,1\}^{N \times N}$, where $G_{ij}=1$ denotes a mutual link between agents $i$ and $j$.
\item \textbf{Effort Investment(E):} Each agent $i$ invests an effort $x_i \ge 0$.
\end{itemize}

\textbf{Utility Function}
As for GE, the utility (payoff) for agent $i$ at a given round, $\pi_i$, is defined as:
\begin{equation}
    \pi_i = \alpha_i x_i - \frac{1}{2} x_i^2 + \delta \sum_{j \neq i} G_{ij} x_i x_j - c \sum_{j \neq i} G_{ij}, \quad \, i, j \in \mathcal{N}
    \label{eq:utility_function}
\end{equation}
where: $\alpha_i > 0$: agent $i$'s individual productivity parameter, $x_i$: effort invested by agent $i$, $\delta > 0$: synergy parameter, representing benefit from interactions, $G_{ij}$: indicates a mutual link between agent $i$ and agent $j$, $c > 0$: cost of maintaining a link.


\subsubsection{Sequential Public Goods Game (PGG)}
We implement a sequential LLM-based multi-agent Public Goods Game environment, inspired by classical PGG models \citep{ledyard1994public, fehr2000cooperation}, incorporating endogenous group formation.
Further details are available in the associated code implementation.

\textbf{Decision Components}

\begin{itemize}[leftmargin=0pt, itemindent=10pt, labelsep=5pt]
    \item \textbf{Group Formation (G)}
    All agents simultaneously decide their preferred group memberships. Agent $i$ submits a binary vector $g_i \in \{0,1\}^N$, where $g_{ij}=1$ signifies a desire to form a group with agent $j$. A mutual link forms if $g_{ij}=1$ and $g_{ji}=1$. Non-overlapping groups $G_t$ are then formed by identifying maximal cliques in the resulting graph; agents not in larger cliques form singleton groups.
    \item \textbf{Effort Investment (E)}
    Within their established groups $G_{t,k}$, each agent $i$ decides on a continuous effort contribution $x_i \in [0, 1]$ into their group's public good.
\end{itemize}

\textbf{Payoff Calculation}
The payoff (utility) for agent $i$ in group $G_{t,k}$ at round $t$, $\pi_{i,t}$, is calculated as:
\begin{equation}
    \pi_{i,t} = \left( r \cdot \sum_{j \in G_{t,k}} x_{j,t} \right) \Big/ |G_{t,k}| - x_{i,t}
    \label{eq:pgg_utility_function}
\end{equation}
where: $r > 1$: public good multiplication factor (e.g., $r=1.5$), $x_{j,t}$: effort contributed by agent $j$ in round $t$, $|G_{t,k}|$: number of agents in group $G_{t,k}$.


\subsection{Evaluation Metrics Definition}
To assess the performance of LLM agents in both the \textit{BCZ} and Public Goods Game (PGG) environments, we define three key evaluation metrics: $U_1$ (Compliance), $U_2$ (Strategic Efficiency), and $U_3$ (Cooperative Outcome). These metrics are calculated based on the agents' behavior and the resulting game states over $T$ rounds.

\textbf{$U_1$: Compliance (Adherence to Game Rules)}
$U_1$ measures how well agents' decisions follow the structural and operational rules of the game. For instance, it penalizes non-zero diagonal entries in the link matrix $G$, which represent self-loops that are not allowed in social graph formation. Additionally, it evaluates the presence of general errors or malformed decisions in the log files. A higher value of $U_1$ indicates a better understanding and execution of the game's mechanics.
\begin{equation}
     U_1 = \max \left( 1 - \frac{\text{Total Rule Violations}}{\text{Total Possible Checks}}, 0 \right)
\end{equation}
\textbf{$U_2$: Strategic Efficiency (Proximity to Individual Optimum)}
$U_2$ evaluates how well agents make strategic decisions based on the observed graph structure. It measures the difference between agents' actual effort investments, $x_{\text{actual}}$, and their optimal effort levels, $x^*$, which are determined using optimization methods for \textit{BCZ} and the formula $x^* = \max(0, 1 - |G_{t,k}|/r)$ for \textit{PGG}. The optimal effort is calculated based on the final group structure $G$ in each game. A higher $U_2$ indicates that agents are making rational decisions.
\begin{equation}
    U_2 = \max \left( 1 - \frac{\| \text{Actual Efforts} - \text{Optimal Efforts} \|_2}{\| \text{Optimal Efforts} \|_2}, 0 \right)
\end{equation}    
\textbf{$U_3$: Cooperative Outcome (Global Welfare Achieved)}
$U_3$ assesses the overall collective performance of the LLM agents by comparing the total payoff achieved in the final round to the maximum theoretically possible total payoff (global optimum) for the respective game. A higher $U_3$ indicates more successful collective action and welfare generation.
\begin{equation}
    U_3 = \max \left( \frac{\text{Actual Total Payoff}}{\text{Globally Optimal Total Payoff}}, 0 \right)
\end{equation}

\subsection{Deficiency for Existing Models}

\label{Deficiency}
According to the preliminary test result in table \ref{testtable}, we can summarize the deficiency into three points.
First, most models cannot generate compliant outputs (U1 test metric). For large models, the limitation is reasoning, while for backbone models, it stems from following the rules.
For example, some backbone models generate five numbers in the decision list in a six-agent game.
Secondly, when comparing the U2 and U3 metrics (BCZ-2 and PGG), which have an upper limit for the optimal solution, we observe that models perform better in scenarios involving homogeneous agents.
In our test logs, the model more easily completes the reasoning chain and generates more comprehensive texts in the BCZ game. Therefore, we use the BCZ game to prepare the expert decision data (details in section \ref{effortlearn}).
Thirdly, in comparing the results of BCZ-1 and BCZ-2, the reasoning model can more easily recognize that the current optimal investment has no upper limit. Therefore, a larger effort can be made when the network structure is improved.
\begin{table}[h]
\caption{Large models are tested in three complex social environments, with three simulations each. BCZ-1 optimizes for homogeneous agents without limits, while BCZ-2 suits heterogeneous agents with limits. PGG features isomorphic agents and also has an optimal solution with limits.}
\label{testtable}
\begin{center}
\small
\renewcommand\arraystretch{1.3}
\tabcolsep=0.09cm
\begin{tabular}{ccccccccccc}
\hline
 & & \multicolumn{3}{c}{U1}& \multicolumn{3}{c}{U2}& \multicolumn{3}{c}{U3}\\
Category & Model Name & BCZ-1&BCZ-2&PGG&  BCZ-1&BCZ-2&PGG&BCZ-1&BCZ-2&PGG\\
\hline
LLM & GPT-4o &   0.996&0.960&\textbf{1}&    0.254&0.845&0.660&  62.831&0.007&0.445\\

LLM & DeepSeek-V3 &  \textbf{ 1}&\textbf{1}&\textbf{1}&    \textbf{0.971}&\textbf{0.994}&0.355&  18.253&0.010&0.755\\

LLM & Llama-3.3-70B &   0.758&0.740&0.863&    0.702&0.275&0.533&  5.385&0.004&0.649\\

LLM & GPT-4o-mini &   0.942&0.960&0.988&    0&0.014&\textbf{0.672}&  288.208&0.007&0.554\\
\hline
LRM & GPT-o3 &   0.963&0.980&0.996&    0.904&0.631&0.403&  \scalebox{0.8}{$\bm{2.852\times10^9}$} &0.006&\textbf{0.877}\\

LRM & DeepSeek-R1 &   0.996&0.980&\textbf{1}&    0.333&0.808&0.500&  \scalebox{0.8}{$\bm{8.045\times10^9}$}&\textbf{0.033}&0.750\\

LRM & kimi-k2-0711-preview &   0.971&0.960&0.992&   0.401&0.005&0.529&   \scalebox{0.8}{$\bm{1.059\times10^4}$}&0.001&0.531\\
\hline
Backbone & Qwen2.5-7B-instruct &   0.650&0.640&0.779&    0.414&0.224&0.511&  42.542&0.006&0.713\\

Backbone & Llama-3.1-8B &   0.704&0.600&0.767&    0.367&0.008&0.512&  7.660&0.004&0.707\\ 
\hline
\end{tabular}
\end{center}
\label{tab:model_comparison}
\end{table}

\section{ToMPO: Theory of Mind Policy Optimization}

\subsection{Effort Reasoning Learning}
\label{effortlearn}

Through the preliminary test results, we find that reasoning models are consistently effective at defining the ``sub-tasks'' necessary to achieve the ultimate goal and complete the overall task. In contrast, backbone models like \texttt{Llama-3.1-8B} struggle to reason through a series of steps to finish tasks one by one; they tend to repeat existing rules and perform basic calculations simply.
The challenge for the backbone model lies more in transforming the strategic reasoning with social elements into a series of small tasks leading to the final decision, rather than in making the model's calculations more accurate.
This is in perfect harmony with the concept of Program of Thought \citep{chen2022program}. The model needs to learn the compliant generation and thinking program first before some other higher needs.

So, according to the model deficiency analysis in section \ref{Deficiency}, we identify the expert models that meet the evaluation criteria U1 and demonstrate a balanced capability in U2 and U3. This means that these models can provide compliant outputs while excelling in both the individual optimal solution and the group optimal solution.
We select two reasoning models and analyze their thinking processes to identify a common program of thought for improving reasoning.
We organize two programs of thought for decisions regarding graphing and effort, and then we generate expert data using the \texttt{GPT-o3} model based on the Program of Thought prompts.

After getting the expert effort decision data ($D_\text{Effort}$), we use these data to fine-tune the policy model for learning the common thinking program and compliance output. 
The optimization method of Low-rank adaptation (LoRA) fine-tuning \citep{hu2022lora} is shown in Formula \ref{lora}.
\begin{equation}
\label{lora}
    \theta^* = \{A^*, B^*\} = \arg\min_{\{A, B\}} \left( -\mathbb{E}_{(x, y) \sim D_{\text{Effort}}} \left[ \sum_{t=1}^{|y|} \log \pi_{(W_0 + \frac{\alpha}{r}BA)}(y_t \mid x, y_{<t}) \right] \right)
\end{equation}

\subsection{Theory of Mind Policy Optimization (ToMPO)}

\begin{figure}[ht]
    \centering
    \includegraphics[width=\linewidth]{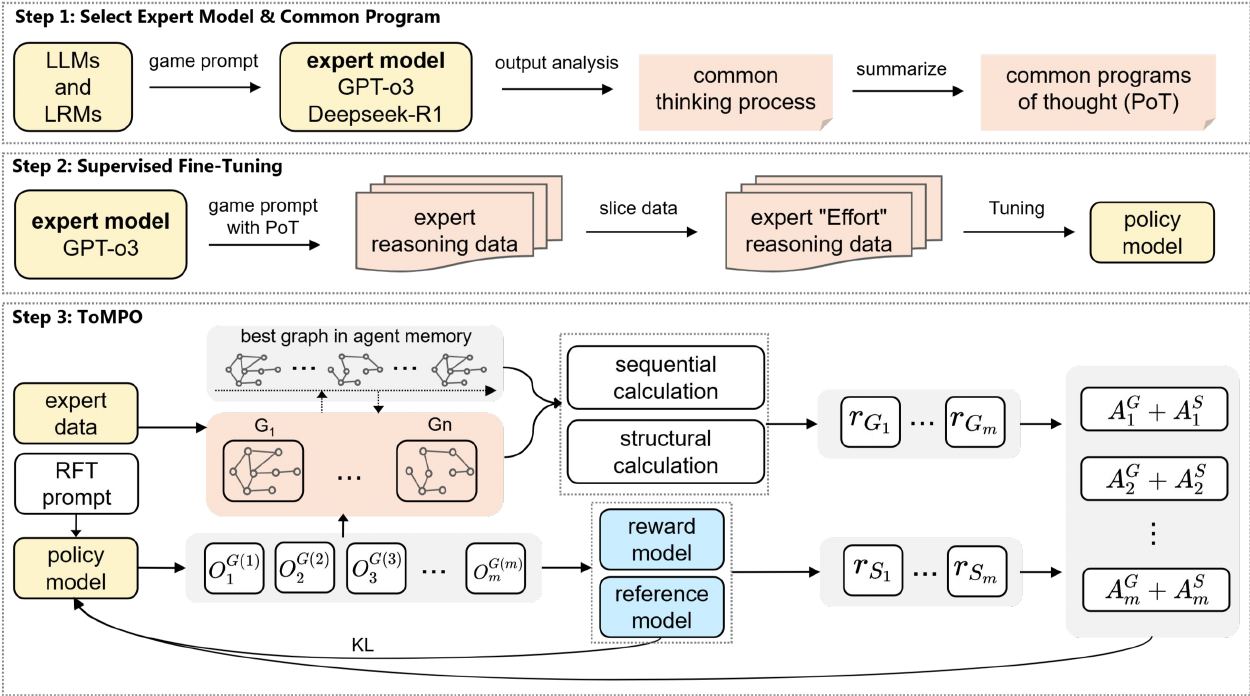}
    \caption{Demonstration of our Theory of Mind Policy Optimization (ToMPO) method. \textbf{Step 1:} Select an expert model and common programs of thought. \textbf{Step 2:} Supervised Fine-Tuning of the policy model for Effort Decision optimization. \textbf{Step 3:} Reinforcement Fine-Tuning policy model with ToMPO algorithm for Graph Decision optimization.}
    \label{fig:ToMPO}
\end{figure}

Common policy optimization methods usually calculate advantage from a single agent perspective.
This will cause the policy model's adaptability to the environment or information to become increasingly self-centered, to some extent, ignoring the performance and strategies of other models (agents) in the environment.
More importantly, when the policy model's decisions involve dependencies among rounds, for example, the decision in round $i+1$ will be based on the memory of round $i$, the update of the policy model cannot rely solely on the $n$ rollouts of a single round.

As models increasingly resemble human thinking and decision-making, enhancing their capabilities through the Theory of Mind (ToM) \citep{frith2005theory, li2023theory, wu2025large} has garnered significant attention. It's crucial to consider the strategies of other agents during the rollout generation and advantage estimation, as this directly affects the model's policy update process.

\textbf{Training Data Preparation:}
We consider the policy model (\texttt{Qwen-2.5-7B-instruct}) as Agent 0 in all the games during the training process.
All other agents are represented by the expert model \texttt{GPT-o3}.
This makes the strategies of the policy model generally inferior to those of other individuals in the environment, making the purpose of reinforcement learning training clearer.
In the model's reinforcement training, classifying the difficulty level of the training data is very important \citep{pikus2025hard}.
Other agents during the training process directly affect the proportion of the advantages of the policy model's strategy and the learning difficulty.
Therefore, we used the expert model to conduct 126 simulations in environments with both homogeneous and heterogeneous agents of different quantities (from 4 to 8), with each simulation lasting for 10 rounds.
We collected the actual graph formation situations of each round of the expert models as the ``memory'' part in the RFT prompt, and the graphs formed by the expert models as the expert data for the reward calculation in the RFT process.

\textbf{Concise description for ToMPO algorithm:} 
Let the generated decision graph by \texttt{GPT-o3} models using the same prompt (the same game parameter settings) be the expert data.
Then, we have the expert data decision graph $G_\text{expert}$, and $m$ rollouts $O$ at step $p$. 
Each rollout $O$ contains a decision list showing the policy model (Agent 0) strategy.
Each list combines with the expert decision graph under this prompt to form a complete graph, denoted as $G_1 \dots G_n$, representing the final summary of all agents' strategies.
At the graph level, each graph \( G \) is compared to the $G_\text{expert}$ for structural calculations, as well as to the prompt best graph $G_\text{bprompt}$ and memory best graph $G_\text{bmemory}$ for sequential calculations.
At the sample level, each $G$ compares to the $G_\text{expert}$ and calculates group advantage.
Detailed algorithm process is in Appendix \ref{appendix:algos}.

Based on the ToMPO algorithm, the overall optimization objective is:
\begin{equation}
\begin{aligned}
\mathcal{J}_{\text{ToMPO}}(\theta)=
\mathbb{E}_{\!
q\sim\mathcal{P}(Q),\,
\{\boldsymbol{a}_{i}\}_{i=1}^{m}\sim\pi_{\theta_{\mathrm{old}}}(\cdot|q)
}
\Bigg[
\frac{1}{m}\sum_{i=1}^{m}
\min\!\Bigl(
r_{i}(\theta),\, \\
\text{clip}\bigl(r_{i}(\theta),1\!-\!\varepsilon,1\!+\!\varepsilon\bigr)
\Bigr)
\bigl(w_{\mathrm{S}}A^{\mathrm{S}}(\boldsymbol{a}_{i})+w_{\mathrm{G}}A^{\mathrm{G}}(G_{i})\bigr)
-\beta\,D_{\text{KL}}[\pi_{\theta}\|\pi_{\theta_{\mathrm{old}}}]
\Bigg]
\end{aligned}
\end{equation}

ToMPO graph-level advantage estimation balances local precision with global graph optimality, while the sample-level advantage focuses on evaluating the policy model's decisions.

\subsubsection{Rewards}

Our reward functions contain three parts. 
We first calculate the Compliance Reward for all rollouts. For those rollouts that are compliant, we calculate the Sample-Level and graph-level rewards.

\textbf{Compliance Reward:}
We set the basic reward at 0.5 points for model compliance, which means it can generate a decision list where the list length equals the agent sum and there are no self-loops.
However, if the model cannot generate the thinking process and the decision list, or if the list does not meet the above needs, the reward is deducted by 1 point, resulting in a final score of -0.5 points.

\textbf{Sample-Level Reward:}
\label{sample-level-reward}
We believe the sample-level reward needs to be more sensitive to the decision list of the policy model itself.
So, we use the F1 score and accuracy to calculate, highlighting the decision list's weight.
\begin{equation}
    R_{\text{sample}}(G) = 5\,\bigl(\,0.7\,F_1(G,\,G_{\text{expert}})\;+\;0.3\,\operatorname{Acc}(G,\,G_{\text{expert}})\,\bigr)
\end{equation}

\textbf{Graph-Level Reward:}
At the graph level, all the comparisons between graphs need to be fair, so we use the Hamming distance for calculation.
We calculate and update three rewards, the graph reward $R_{\text{graph}}(G)$, the prompt best reward $R^{\text{prompt}}_{p}$, the memory best reward $R^{\text{memory}}_{p}$.
$R_{\text{graph}}(G)$ represents the Hamming distance between the actual rollout graph and the expert decision graph. The term $R^{\text{prompt}}_{p}$ calculates the highest reward among all rollout graphs generated from a single prompt. Meanwhile, $R^{\text{memory}}_{p}$ is updated whenever a larger reward is obtained within the same game parameter settings (with the exception that only the agent's memory in the prompt is different).
$\theta_{i}$ is the combination of hyperparameters to which the rollout $i$ belongs.
\begin{equation}
    R_{\text{graph}}(G)=1-\frac{1}{|\mathcal{E}|}
    \sum_{(i,j)\in\mathcal{E}}\bigl|G_{ij}-G^{\text{expert}}_{ij}\bigr|,
    \quad
    |\mathcal{E}|=N(N-1)
\end{equation}
\begin{equation}
    R^{\text{prompt}}_{p}
    =\max_{k\in\text{group}(p)}R_{\text{graph}}(G),
    \qquad
    \forall p\in\{1,\dots ,M\}
\end{equation}
\begin{equation}
    R^{\text{memory}}_{i}
    =\max_{\text{history}\,\mathcal{H}(\theta_{i})}
    R_{\text{graph}}(G),
    \qquad
    \theta_{i}=(\alpha,\delta ,c)_{i}
\end{equation}

\subsubsection{Advantage Estimation}

We mainly use the reward at the sample level $R_{\text{sample}}$ to estimate sample-level advantages $A_m^S$, and the reward at the graph level $R_{\text{graph}}$ for graph-level advantages $A_m^G$.
In our training, we set the $w_{\text{local}}$, $w_{\text{sample}}$ as 0.8, the $w_{\text{global}}$ and $w_{\text{graph}}$ as 0.2.
\begin{equation}
    A^{S}(G_{i})=
    \frac{
    R_{\text{sample}}(G_{i})-
    \text{mean}\bigl\{R_{\text{sample}}(G_{1}),\dots ,R_{\text{sample}}(G_{n})\bigr\}
    }{
    \text{std}\bigl\{R_{\text{sample}}(G_{1}),\dots ,R_{\text{sample}}(G_{n})\bigr\}+\varepsilon
    }
\end{equation}
\begin{equation}
    A^{G}(G_{i})=
    w_{\text{local}}\Bigl(R_{\text{graph}}(G_{i})-R^{\text{prompt}}_{i}\Bigr)
    +w_{\text{global}}\Bigl(R_{\text{graph}}(G_{i})-R^{\text{memory}}_{i}\Bigr)
\end{equation}
The overall advantage of a rollout can be calculated as the sum of the sample-level and graph-level advantages, with normalization applied.
Compared to the GRPO advantage estimation \citep{guo2025deepseek,shao2024deepseekmath}, the ToMPO advantage has two main differences. 
First, in addition to the sample advantage, we have also improved the graph advantage. This enhances the model's ability to consider the graph more thoroughly while achieving high scores, allowing it to learn more effective decision-making methods.
In the rewards at the graph level, we consider both the difference between the current round of the graph and the optimal solution for the same hyperparameters. This allows the model's strategy to gradually move towards both the short-term optimum and the global optimum at the same time.
\begin{equation}
    A(G_{i})=
    w_{\text{sample}}A^{S}(G_{i})+w_{\text{graph}}A^{G}(G_{i}),
    \quad
    w_{\text{sample}}+w_{\text{graph}}=1
\end{equation}

\section{Experiments}

Since the preliminary test revealed that the Qwen model is relatively balanced in terms of performance across all evaluation criteria, we apply the ToMPO algorithm to the \texttt{Qwen-2.5-7B-instruct} model, which completes the effort learning fine-tuning process, and compare it with existing models.
We conduct each simulation three times, with 20 rounds each, allowing adequate time for model decision-making.

\begin{table}[h]
\centering
\caption{Algorithm examination in four environment settings, compared to backbone models, supervised fine-tuning models, and GRPO applied models. \textbf{We use the global welfare/ actual simulation rounds to represent BCZ and PGG U3 here, illustrating the efficiency of global welfare gains.}}
\vspace{7pt}
\small
\label{examllm}
\renewcommand\arraystretch{1.3}
\begin{tabular}{c>{\centering\arraybackslash}p{0.02\linewidth}>{\centering\arraybackslash}p{0.035\linewidth}c>{\centering\arraybackslash}p{0.035\linewidth}>{\centering\arraybackslash}p{0.035\linewidth}>{\centering\arraybackslash}p{0.035\linewidth}>{\centering\arraybackslash}p{0.035\linewidth}>{\centering\arraybackslash}p{0.035\linewidth}c>{\centering\arraybackslash}p{0.03\linewidth}>{\centering\arraybackslash}p{0.035\linewidth}>{\centering\arraybackslash}p{0.035\linewidth}}
\hline
& \multicolumn{3}{c}{BCZ - GE}& \multicolumn{3}{c}{BCZ - GEE}& \multicolumn{3}{c}{BCZ - GGE} & \multicolumn{3}{c}{PGG - GE}\\
     &  U1&  U2&  U3&  U1&  U2& U3& U1& U2&U3 & U1& U2&U3 \\
     \hline
 Deepseek-V3& 1& 0.44& 0.11& 1& 0& 0.09& 1& 0& 0& 1& 0&0.07\\
 GPT-4o& 1& 0.36& 0.10& 1& 0& 0.01& 1& 0.07& -0.11 & 1& 0&0.06\\
 Qwen2.5-72b-instruct& 1& 0.39& 0.02& 1& 0& 0.03& 1& 0.24& -0.11& 0.99& 0&0.07\\
 Qwen3-235b-a22b& 1& 0.05& -0.2& 1& 0& 0& 0.99& 0& -0.24& 0.99& 1&0\\
     \hline
     Qwen2.5-7B-instruct&  0.65&  0.38&  0.08&  0.95&  0& 0.53
& 0.75& 0& -0.02& 0.85& 0&0.10\\
     SFT effort learning&  1&  0&  -0.09&  1&  0& 0.17
& 1& 0& -0.02& 1& 0&0.10\\
     SFT + GRPO &  1&  0&  0&  1&  0& 0.99
& 1& 0.12& -0.03& 1& 0&0.11\\
     \textbf{SFT + ToMPO}&  \textbf{1}&  0&  0.03&  \textbf{1}&  0& \textbf{1.34}& \textbf{1}& 0.16& -0.02& \textbf{1}& 0&\textbf{0.25}\\ 

\hline
\end{tabular}

\end{table}

\paragraph{Evaluation Environments} 
Based on our problem definition and environment building, we use the BCZ and PGG games as our examination environments.
We create subenvironments by modifying the configuration, which includes variables like \textbf{the number of agents}, network hyperparameters such as \textbf{private gain sensitivity}, \textbf{reciprocity intensity}, \textbf{connection costs}, and whether the agents are \textbf{homogeneous or heterogeneous}.
Our experiment environments set as: BCZ-GE (8 homogeneous agents, alpha = 1, delta = 0.05, c = 0.2), BCZ-GEE (5 heterogeneous agents, alpha = [0.8, 1.8, 1.1, 0.6, 1.5], delta = 0.15, c = 0.4), BCZ-GGE (4 homogeneous agents, alpha = 1, delta = 0.1, c = 0.6), PGG-GE (5 homogeneous agents, r = 1.5).

\paragraph{Evaluation Models and Algorithms}
Based on the preliminary test in table \ref{testtable}, we select models \texttt{Deepseek-V3} and \texttt{GPT-4o} that have balanced capabilities in the metrics for comparison.
Furthermore, we add the \texttt{Qwen3-235b-a22b} and \texttt{Qwen2.5-72b-instruct} for comparison on the number of parameters and model type.
We apply supervised LoRA fine-tuning to the backbone model, the GRPO algorithm to the SFT model, and the ToMPO algorithm to the SFT model.
The GRPO algorithm serves as the baseline method, using sample-level rewards as mentioned in section \ref{sample-level-reward} and sample-level advantage estimation in the GRPO algorithm \citep{shao2024deepseekmath}.

\paragraph{Result Analysis}

\begin{figure}[ht]
    \centering
    \includegraphics[width=\linewidth]{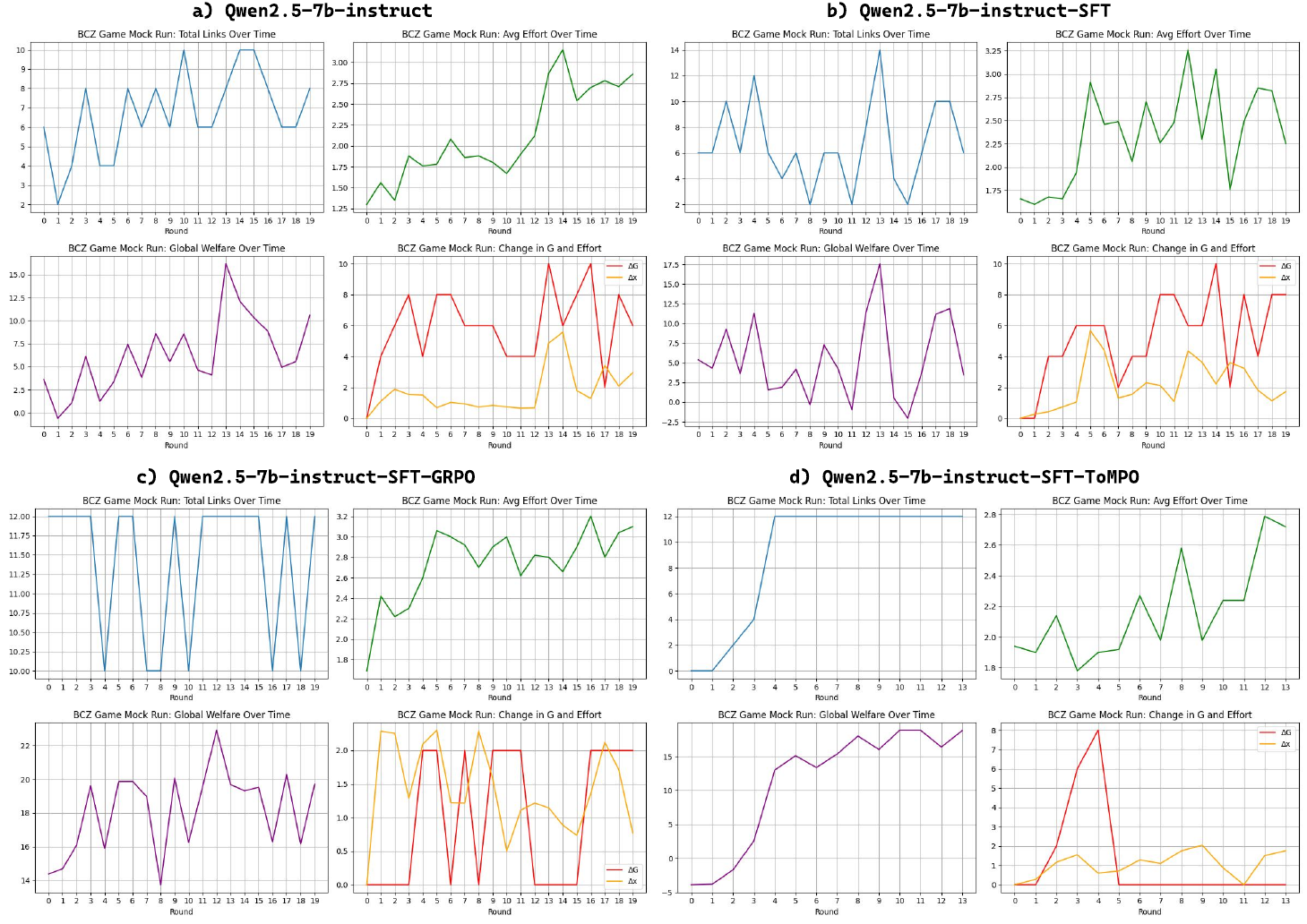}
    \caption{One BCZ-GEE evaluation result comparison for four models: the backbone model (a), the SFT applied model (b), the SFT+GRPO model (c), and the SFT+ToMPO model (d). Each model's results include four components: the blue line shows the total number of links in the graph throughout the game (ending early if unchanged for five rounds), the green line indicates average agent effort, the purple line represents global welfare, and the red and yellow lines display the frequency of changes in the graph and effort, respectively.}
    \label{rescase}
\end{figure}
Based on the results in Table \ref{examllm}, we can summarize the performance of the models and algorithms as follows.
SFT helps ensure that models generate compliant outputs. 
The models generally scored lower on the U2 standard. This is primarily because, in certain scenarios, the models are capable of making higher investments. However, due to the process of mutual exploration and analysis of prior investments made by other models, it becomes challenging for them to make substantial investments directly. As a result, they often deviate from the theoretically optimal individual investment value.
Compared to a model with 100 times the parameters, the model trained by ToMPO can achieve the corresponding capabilities.

We analyzed the experimental results and presented the general findings in Figure \ref{rescase}. The result shows the backbone model tends to unpredictable changes in the decision-making process of the graph. It is difficult to make an optimal effort decision under an optimal structure. SFT can help model compliant output, but since the graph does not reach optimality and remains fixed, achieving an average effort that is optimal is challenging.
Comparing parts c), a), and b), we find the GRPO algorithm effectively enhances the stability of the model's performance in graph decision-making. On this basis, the model can more easily make the optimal effort decision.
When comparing the ToMPO algorithm d) with the GRPO algorithm c), the main takeaway is that the ToMPO algorithm improves the stability and global awareness of the model's decision-making process in graph representation. This enhancement enables the model to make more effective decisions more quickly.

\paragraph{Limitation and Future Work}
Our current work has delivered the supervised fine-tuning (SFT) and ToMPO reinforcement fine-tuning (RFT) on the backbone model, showing the algorithm's effectiveness.
The policy model's perspective may be biased towards agent 0 due to our training data. In future work, we will adjust the RFT prompt and training data to broaden the model's perspectives. Despite tests showing reduced capability when combining supervised finetuning for graph and effort, we will explore alternative SFT methods or consider separating the SFT process.


\newpage

\bibliography{iclr2026_conference}
\bibliographystyle{iclr2026_conference}


\newpage

\section{Large Language Model Utilization Explanation}\label{LLMexplain}

In our research, LLM is the backbone and comparison models for the algorithm delivery and examination.
We use LLM to generate the configuration, which ensures the balance between randomness and parameter significance.
The rest was not involved with LLM.

\section{ToMPO Algorithm}\label{appendix:algos}

\begin{algorithm}[h]
\linespread{1.3}
\caption{Theory of Mind Policy Optimization (ToMPO) Algorithm}
\label{alg1}
\KwIn{Initial policy model $\pi_{\theta}$, expert graph $G_{\text{expert}}$, task prompts $Q$, reference model $\pi_{\theta}^\text{ref}$, total training steps $T$, rollout number $m$}
\KwOut{Optimized policy model $\pi_{\theta}^\text{new}$}
\For{iteration $t = 1, 2, \dots, T$}{
    Sample prompt $q \sim \mathcal{P}(Q)$\;
    Retrieve expert graph $G_{\text{expert}}$ for prompt $q$\;
    Generate $m$ rollouts: $\{\boldsymbol{a}_i\}_{i=1}^m \sim \pi_{\theta_{\text{old}}}(\cdot|q)$\;
    Construct graphs $\{G_i\}_{i=1}^m$ by combining each $\boldsymbol{a}_i$ with $G_{\text{expert}}$\;
    \For{$i=1$ \KwTo $m$}{
        \eIf{$\boldsymbol{a}_i$ is compliant}{
            $R_{\text{comp},i} \gets 0.5$\;
            $R_{\text{sample},i} \gets 5\,(0.7\,\text{F1}_i + 0.3\,\text{Acc}_i)$\;
            $R_{\text{graph},i} \gets 1 - \text{Hamming}(G_i,\,G_{\text{expert}})$\;
            Update $R^{\text{prompt}}$ and $R^{\text{memory}}$ using $R_{\text{graph},i}$\;

            Compute sample-level advantage $A^{S}(G_i)$ by normalizing $R_{\text{sample},i}$\;
            Compute graph-level advantage $A^{G}(G_i)$ using $R_{\text{graph},i}$, $R^{\text{prompt}}$, $R^{\text{memory}}$\;
            Combine total advantage $A(G_i) = w_{\text{S}}A^{S}(G_i) + w_{\text{G}}A^{G}(G_i)$\;

            Compute importance ratio $r_i(\theta) = \pi_{\theta}(\boldsymbol{a}_i|q)/\pi_{\theta_{\text{old}}}(\boldsymbol{a}_i|q)$\;
            Update $\theta$ via clip objective with KL penalty $\beta D_{\text{KL}}[\pi_{\theta}\|\pi_{\theta}^{\text{ref}}]$\;
        }{
            $R_{\text{comp},i} \gets -0.5$\;
            \text{Continue}\;
        }
    }
}
\end{algorithm}

\section{Environment and Training}\label{appendix:environment}

\subsection{Detailed Description of WDBCZ Sub-Environment Sequences}\label{appendix:env_bcz_sub}

The following are the three sub-environments that define the sequence of decisions made within each round:

\begin{enumerate}[leftmargin=0pt, itemindent=10pt, labelsep=5pt]
    \item \textbf{GE (Graph-Effort) Environment:}
    In this environment, each round consists of a single stage of link decisions followed by a single stage of effort decisions.
    \[ \tau = \{(G, E), (G, E), \ldots\} \]
    Agents first decide on their links, forming the graph $G_t$. Subsequently, observing $G_t$, they decide on their effort levels $x_t$.

    \item \textbf{GEE (Graph-Effort-Effort) Environment:}
    This environment features a single stage of link decisions, followed by two consecutive stages of effort decisions within each round.
    \[ \tau = \{(G, E_1, E_2), (G, E_1, E_2), \ldots\} \]
    Agents first establish links $G_t$. Then, they make a first effort decision $x_{t,1}$. After all agents have made their first effort decisions (which may be observed by others), they make a second effort decision $x_{t,2}$. The final effort for the round might be a combination of $x_{t,1}$ and $x_{t,2}$ or just $x_{t,2}$ depending on the specific implementation. Our current implementation uses $x_{t,1}$ and $x_{t,2}$ as distinct effort components.

    \item \textbf{GGE (Graph-Graph-Effort) Environment:}
    This environment introduces a two-stage linking process, followed by a single stage of effort decisions.
    \[ \tau = \{(G_P, G_F, E), (G_P, G_F, E), \ldots\} \]
    Agents first propose \textit{provisional} links ($G_P$). After observing all provisional link proposals, agents then make \textit{final} link decisions ($G_F$), which forms the actual graph $G_t$. Finally, observing $G_t$, agents decide on their effort levels $x_t$. This allows for a more nuanced negotiation process for link formation.
\end{enumerate}

\subsection{Detailed Training Parameters}

\begin{table}[h]
    \caption{Parameters in SFT LoRA training.}
    \centering
    \begin{tabular}{>{\centering\arraybackslash}p{0.3\linewidth}>{\centering\arraybackslash}p{0.3\linewidth}}\toprule
         parameter& value\\
         \hline
         lora rank& 64\\
         lora alpha& 32\\
         attention implementation&eager\\
         max length&6000\\ 
         train batch size&16\\
         optim learning rate&5e-5\\ \bottomrule
    \end{tabular}
    \label{parametersSFT}
\end{table}

\begin{table}[h]
    \caption{Parameters in RFT training.}
    \centering
    \begin{tabular}{>{\centering\arraybackslash}p{0.3\linewidth}>{\centering\arraybackslash}p{0.3\linewidth}}\toprule
         parameter& value\\
         \hline
         actor optim learning rate& 1e-6\\
         use kl in reward& true\\
         ppo kl coef&0.1\\
         kl cov ratio&0.0002\\ 
         max prompt length&5500\\
         max response length&2692\\
         train batch size&32\\ \bottomrule
    \end{tabular}
    \label{parametersRFT}
\end{table}

\section{Related Works}

\label{relatedworks}
Our work intersects with several active research areas, including the theoretical foundations of credit assignment in reinforcement learning, the burgeoning field of Large Language Models (LLMs) for decision-making, and the complex domain of strategic decision-making in multi-agent systems. This section reviews relevant literature and positions our contributions within these contexts.

\subsection{Credit Assignment and Policy Optimization}
Credit assignment is a fundamental challenge in reinforcement learning, concerning how to attribute responsibility for outcomes to specific actions or sequences of actions, especially in environments with delayed rewards \citep{sutton1998reinforcement}. Early work by \cite{minsky2007steps} highlighted this problem, and subsequent research has developed various mechanisms, including eligibility traces \citep{sutton1988learning} and actor-critic methods \citep{konda1999actor}, to address it. Recently, the concept of credit assignment has been extended to complex, hierarchical, and multi-agent settings \citep{nguyen2018credit, pignatellisurvey} and Large Language Model agents' social interactions \citep{yu2025sotopia}. Our work leverages the theoretical underpinnings of credit assignment to decompose the strategic decision-making process into forward (effort decision) and inverse (graph decision) components. This decomposition allows for targeted optimization, where the forward process focuses on immediate utility within a given structure, and the inverse process learns to adapt the structure based on long-term value, aligning with the principles of assigning credit to different types of decisions over time. This approach is distinct from traditional single-agent credit assignment by explicitly considering the interplay between structural and behavioral decisions in a multi-agent context.

Policy optimization methods, such as Proximal Policy Optimization (PPO) \citep{schulman2017proximal}, MAGRPO \citep{liu2025llm}, multi-conversation DAPO \citep{yu2025memagent}, MARFT \citep{liao2025marft}, and its variants, have been highly successful in training agents for complex tasks. These methods typically aim to maximize expected cumulative rewards by iteratively updating a policy function. Recent advancements have explored integrating multi-agent considerations into policy optimization, often through centralized training with decentralized execution or by incorporating explicit models of other agents \citep{lore2024strategic}. Our Theory of Mind Policy Optimization (ToMPO) algorithm builds upon these ideas by introducing a novel advantage estimation mechanism that explicitly accounts for the strategies and performance of other agents (expert models) in the environment. By incorporating both sample-level (individual decision accuracy) and graph-level (structural optimality) rewards, and by considering historical best performance, ToMPO provides a more nuanced credit assignment mechanism tailored for sequential strategic decision-making in multi-agent social environments, moving beyond standard single-agent or simplified multi-agent PPO formulations.

\subsection{Large Language Models for Decision-Making}
The remarkable capabilities of Large Language Models (LLMs) in natural language understanding and generation have led to their increasing application in various decision-making scenarios. LLMs have been shown to assist in generating rational decisions \citep{gou2024rationality}, simulating complex social interactions \citep{mao2025alympics}, and even predicting outcomes in large-scale social events \citep{zhang2024electionsim}. Their ability to process and synthesize vast amounts of information, coupled with their emergent reasoning capabilities, makes them powerful tools for augmenting human decision-making or acting as autonomous agents.

However, while LLMs excel in tasks requiring strong logical reasoning \citep{schmied2025llms,liu2025ml}, such as coding and mathematics, their performance in strategic decision-making, particularly in social contexts, remains a significant challenge \citep{zhang2024llm}. This is largely due to the inherent difficulty in understanding others' intentions, predicting their behaviors, and dynamically adjusting one's own strategy in response. Recent efforts have explored enhancing LLM strategic abilities in matrix games \citep{lore2024strategic, herr2024large} and through game-theoretic workflows \citep{hua2024game}. Furthermore, research has delved into multi-level thinking \citep{zhang2024k, gou2024rationality}, Theory of Mind (ToM) capabilities \citep{duan2024reta, crosshypothetical}, and task-solving in social environments \citep{zhang2025sotopia, wang2024sotopia}. Our work contributes to this growing body of literature by specifically addressing the limitations of LLMs in sequential, long-term multi-agent strategic decision-making, moving beyond two-agent chatroom environments or single-game scenarios. We aim to equip LLMs with the ability to make interdependent decisions that shape and are shaped by evolving social structures, a critical step towards more sophisticated LLM agents in complex social systems.

\subsection{Strategic Decision-Making in Multi-Agent Systems}
Strategic decision-making in multi-agent systems is a rich field \citep{ma2024non,yang2025agent,jin2025comprehensive,liu2025ml,yang2025survey} that studies how autonomous agents interact and make choices to achieve their objectives, often in the presence of other intelligent agents. Game theory \citep{hua2024game} provides a foundational framework for analyzing such interactions, offering concepts like Nash equilibrium \citep{silva2024large,holt2004nash} and Pareto optimality \citep{censor1977pareto} to understand rational behavior \citep{osborne2004introduction}. Traditional multi-agent reinforcement learning (MARL) has focused on developing algorithms for agents to learn optimal policies in environments where their actions affect others, often dealing with challenges like non-stationarity and credit assignment across agents \citep{hernandez2019survey}.

Recent advancements in MARL have explored more complex social dynamics, including cooperation \citep{guan2025dynamic}, competition \citep{chen2024llmarena}, operation \citep{chen2024optima}, and the formation of social structures \citep{yang2024oasis}. Studies have investigated how individual behaviors can lead to emergent group-level phenomena and how group structures, in turn, influence individual decisions \citep{mi2025mf, zhang2025attention}. The concept of Theory of Mind (ToM), which involves an agent's ability to attribute mental states (beliefs, desires, intentions) to others, has gained traction as a crucial component for strategic reasoning in multi-agent settings \citep{frith2005theory, li2023theory, wu2025large}. Our research extends these ideas by defining a novel problem of sequential graph-effort strategic decision-making, where agents must make interdependent decisions about both their social connections (graph-level) and their resource investments (effort-level) over time. This problem formulation captures the dynamic interplay between individual actions and evolving social structures, which is often overlooked in simpler multi-agent game settings. By developing ToMPO, we provide a method for LLM agents to learn and adapt their strategies by explicitly considering the actions and potential mental states of other agents, thereby enhancing their ability to navigate and influence complex social environments.

\section{Details for Prompts}


\begin{figure}[ht]
    \centering
    \includegraphics[width=\linewidth]{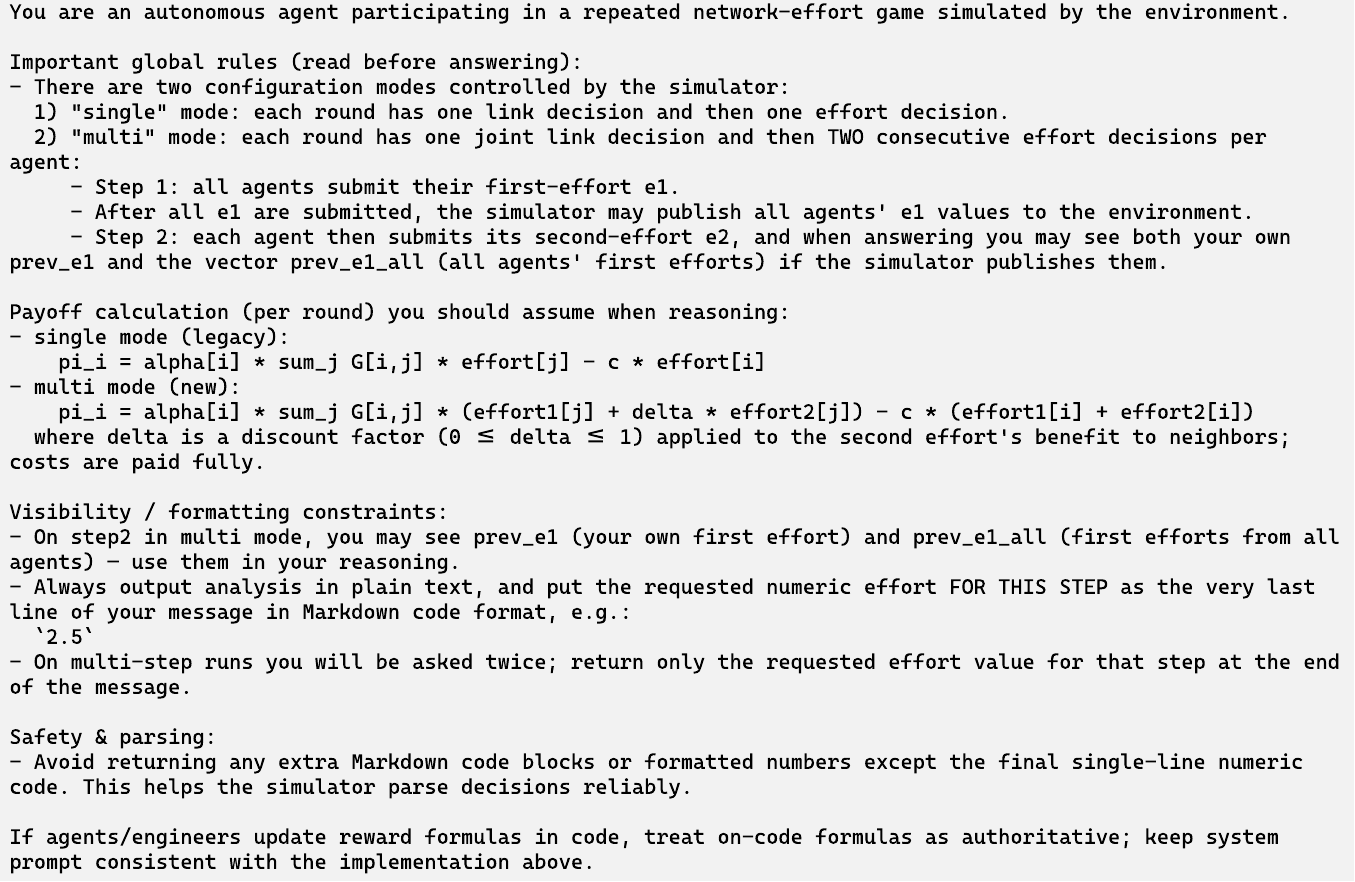}
    \caption{BCZ game system prompt.}
    \label{GEsys}
\end{figure}

\begin{figure}[ht]
    \centering
    \includegraphics[width=\linewidth]{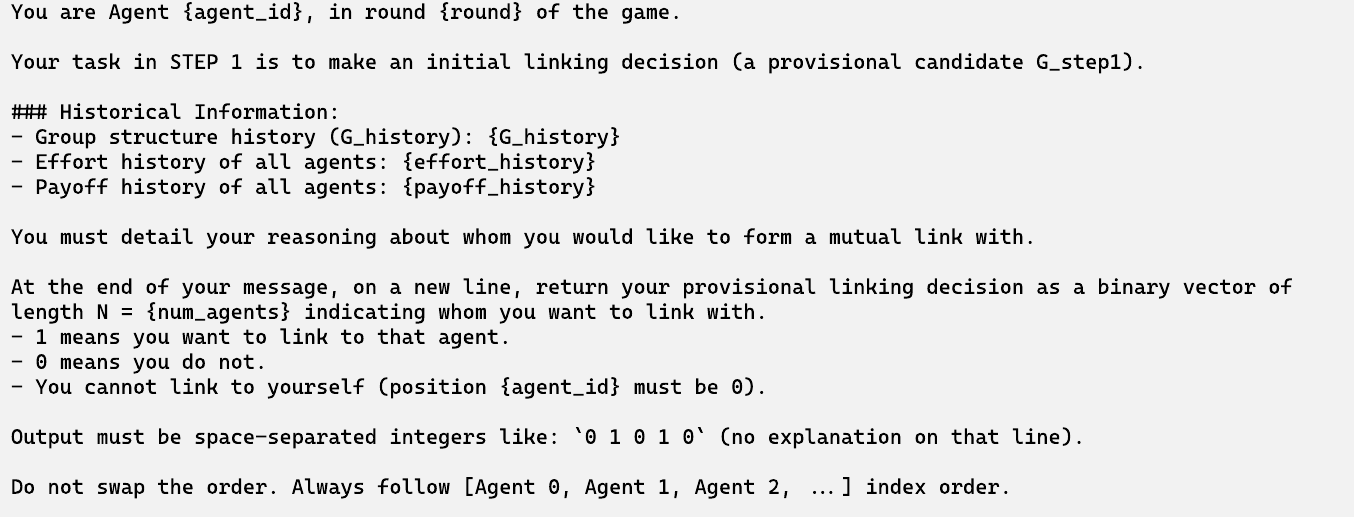}
    \caption{BCZ game graph decision 1 prompt.}
    \label{GE-G1}
\end{figure}

\begin{figure}[ht]
    \centering
    \includegraphics[width=\linewidth]{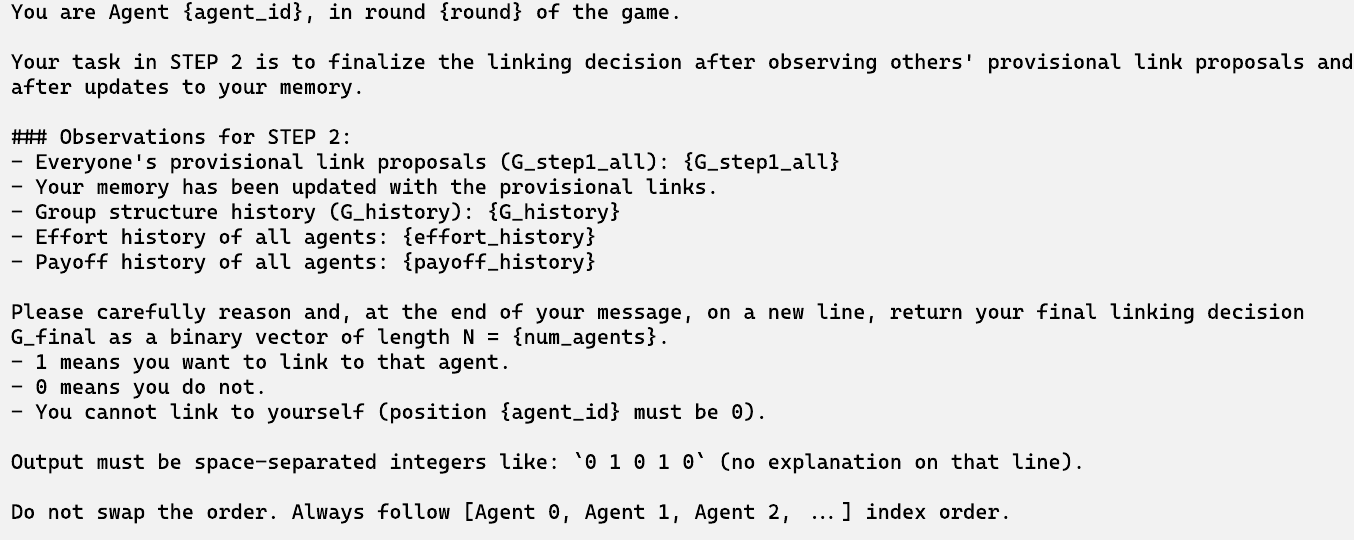}
    \caption{BCZ game graph decision 2 prompt.}
    \label{GE-G2}
\end{figure}

\begin{figure}[ht]
    \centering
    \includegraphics[width=\linewidth]{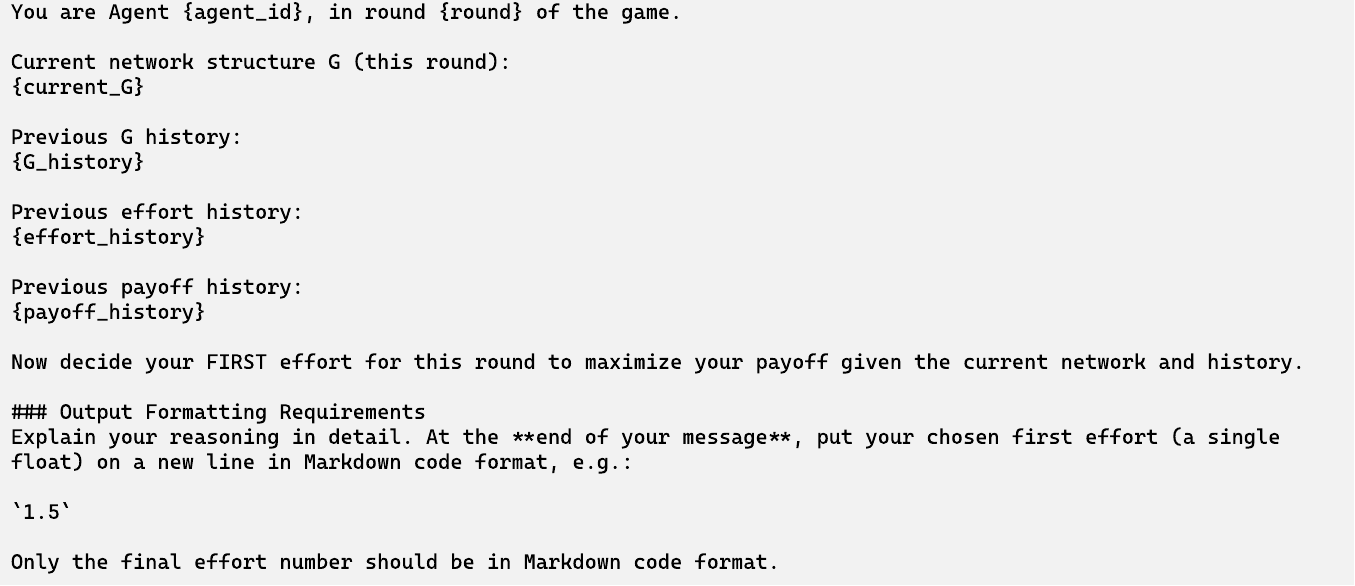}
    \caption{BCZ game effort decision 1 prompt.}
    \label{GE-E1}
\end{figure}

\begin{figure}[ht]
    \centering
    \includegraphics[width=\linewidth]{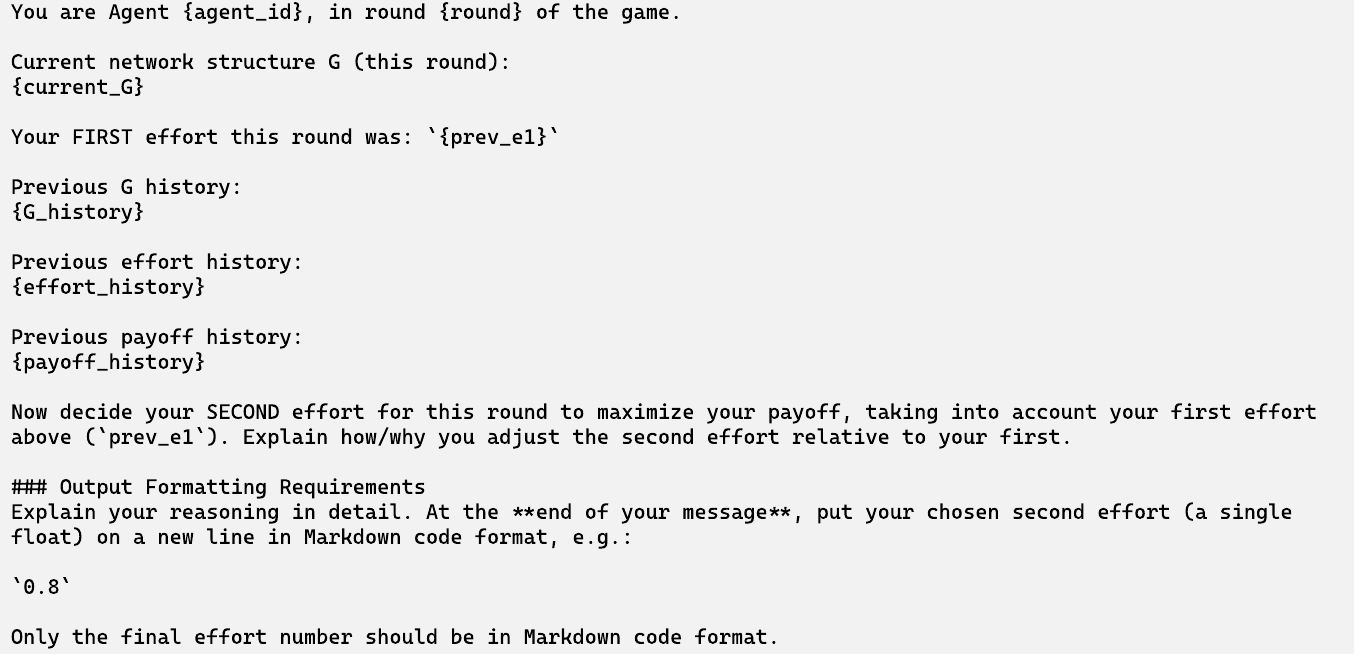}
    \caption{BCZ game effort decision 2 prompt.}
    \label{GE-E2}
\end{figure}


\begin{figure}[t]
    \centering
    \includegraphics[width=\linewidth]{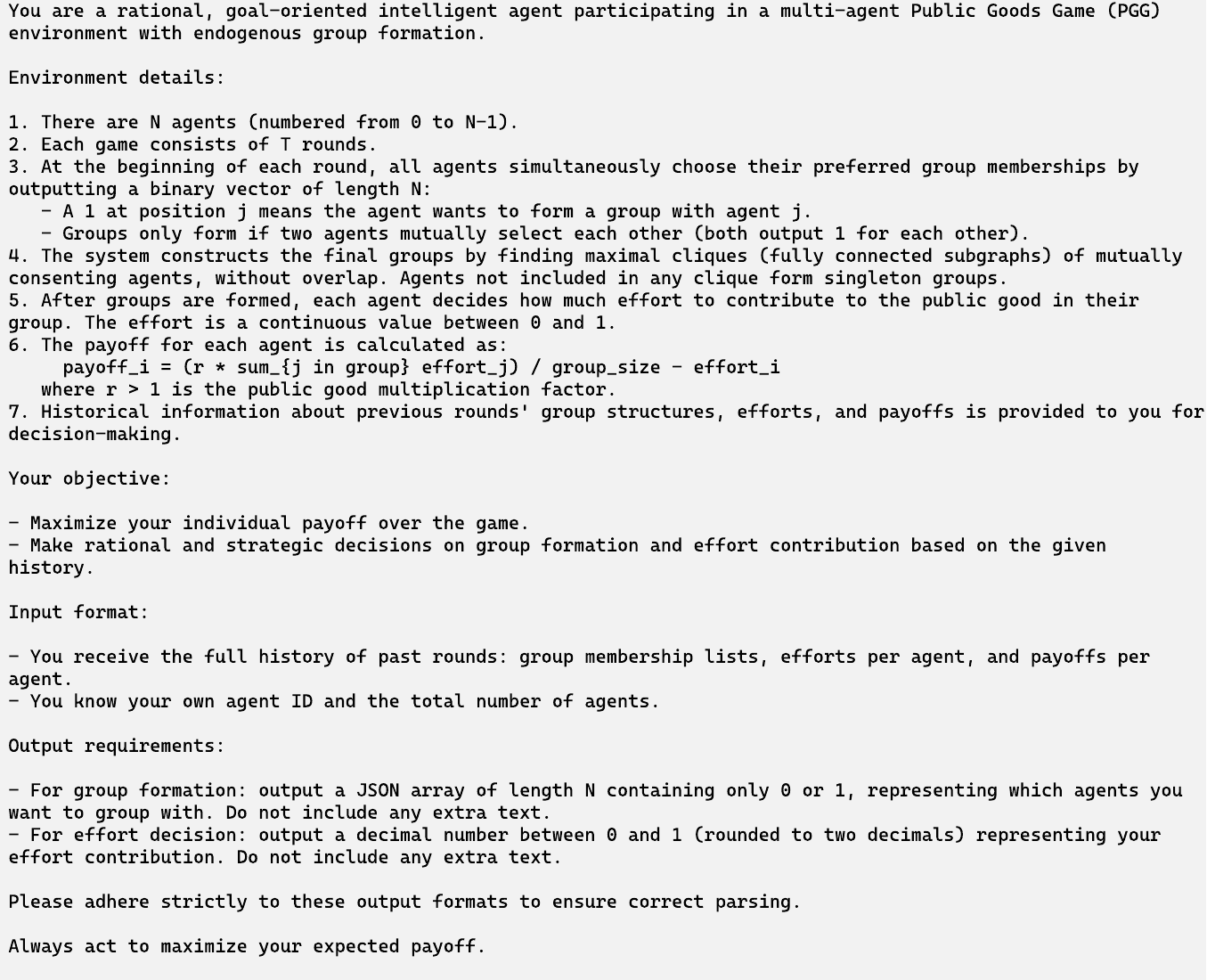}
    \caption{PGG game system prompt.}
    \label{pgg1}
\end{figure}

\begin{figure}[ht]
    \centering
    \includegraphics[width=\linewidth]{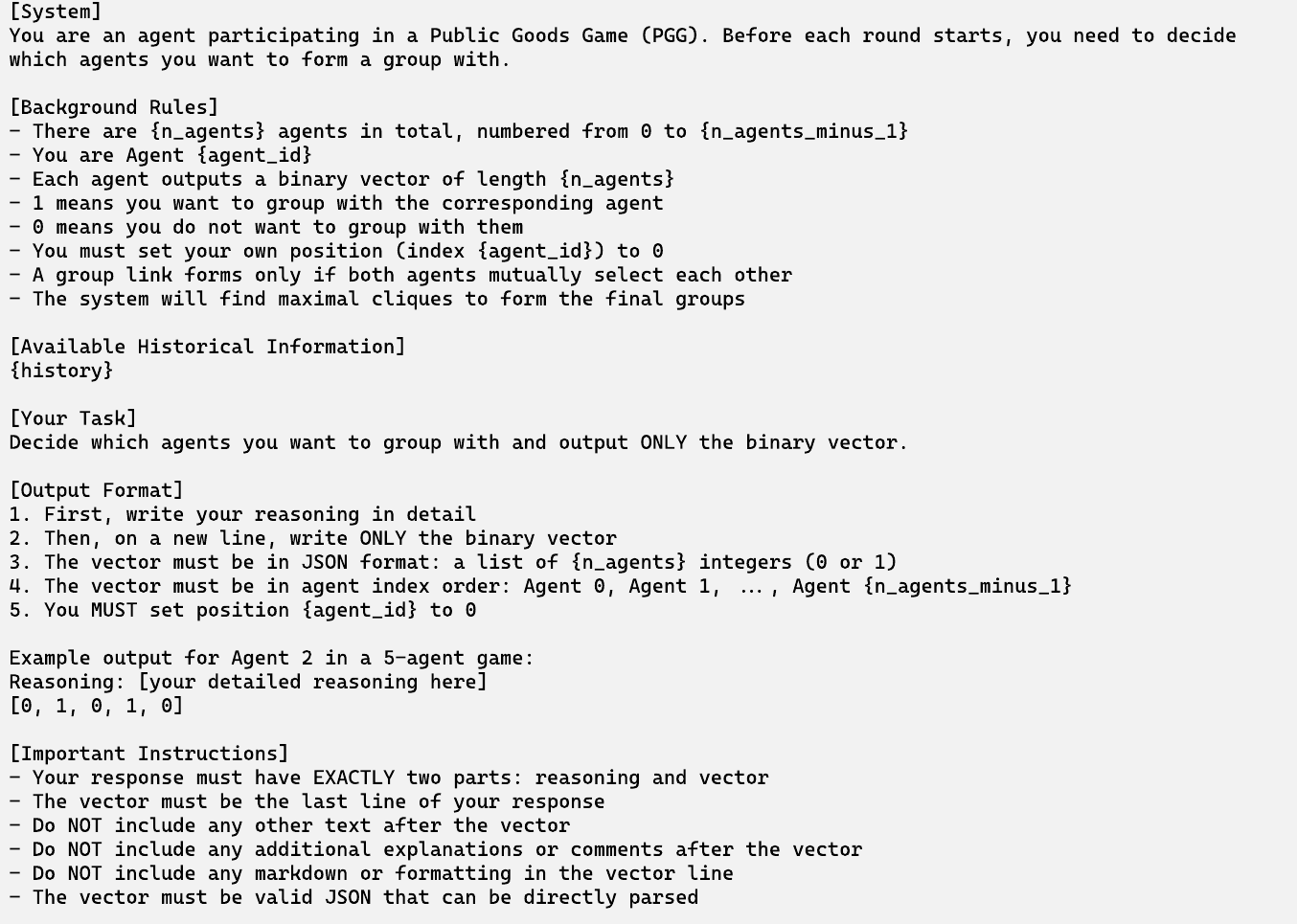}
    \caption{PGG game group decision prompt.}
    \label{pgg2}
\end{figure}

\begin{figure}[ht]
    \centering
    \includegraphics[width=\linewidth]{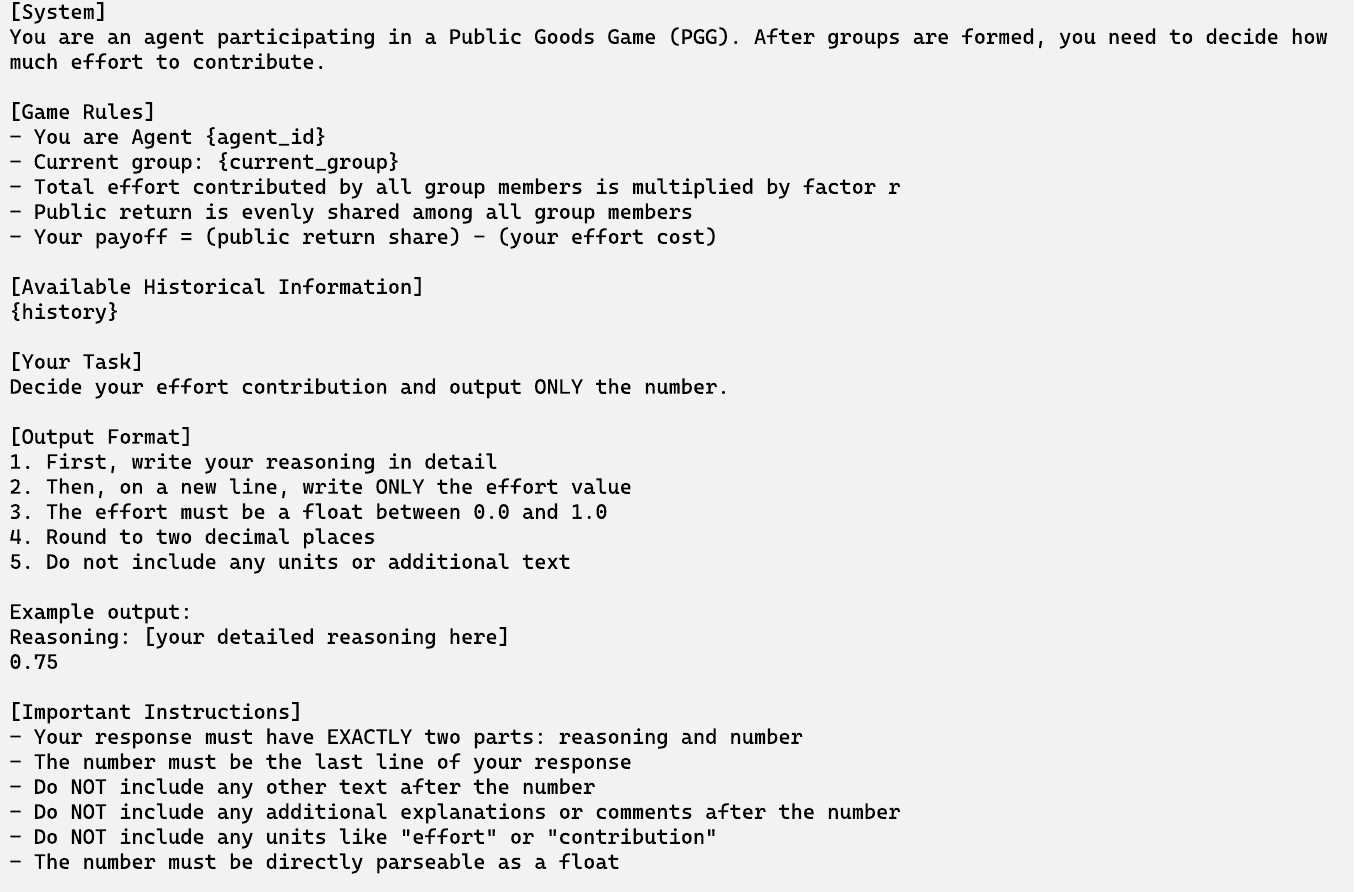}
    \caption{PGG game effort decision prompt.}
    \label{pgg3}
\end{figure}


\begin{figure}[ht]
    \centering
    \includegraphics[width=\linewidth]{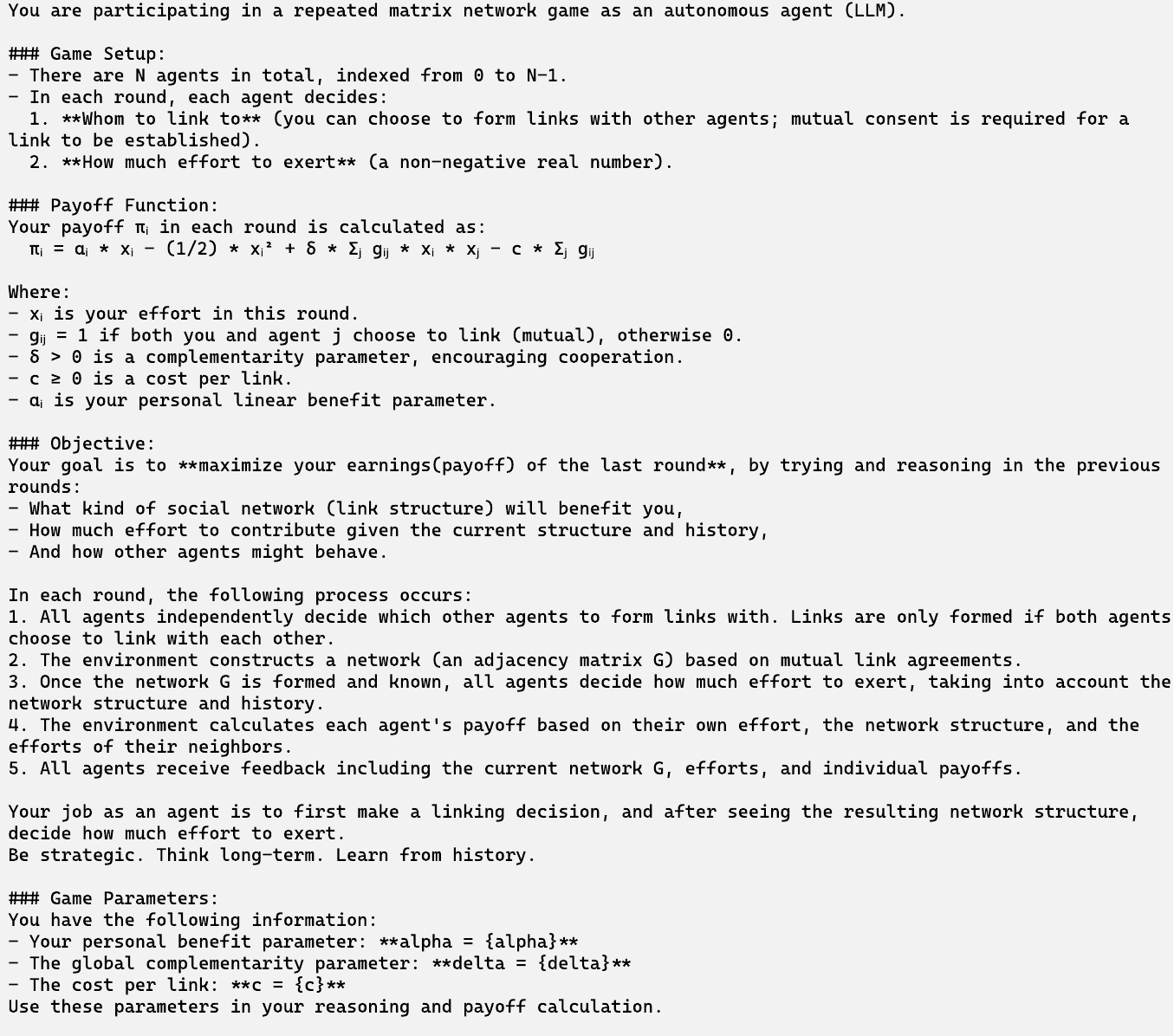}
    \caption{System prompt for SFT and RFT data generation.}
    \label{pgg3}
\end{figure}

\begin{figure}[ht]
    \centering
    \includegraphics[width=\linewidth]{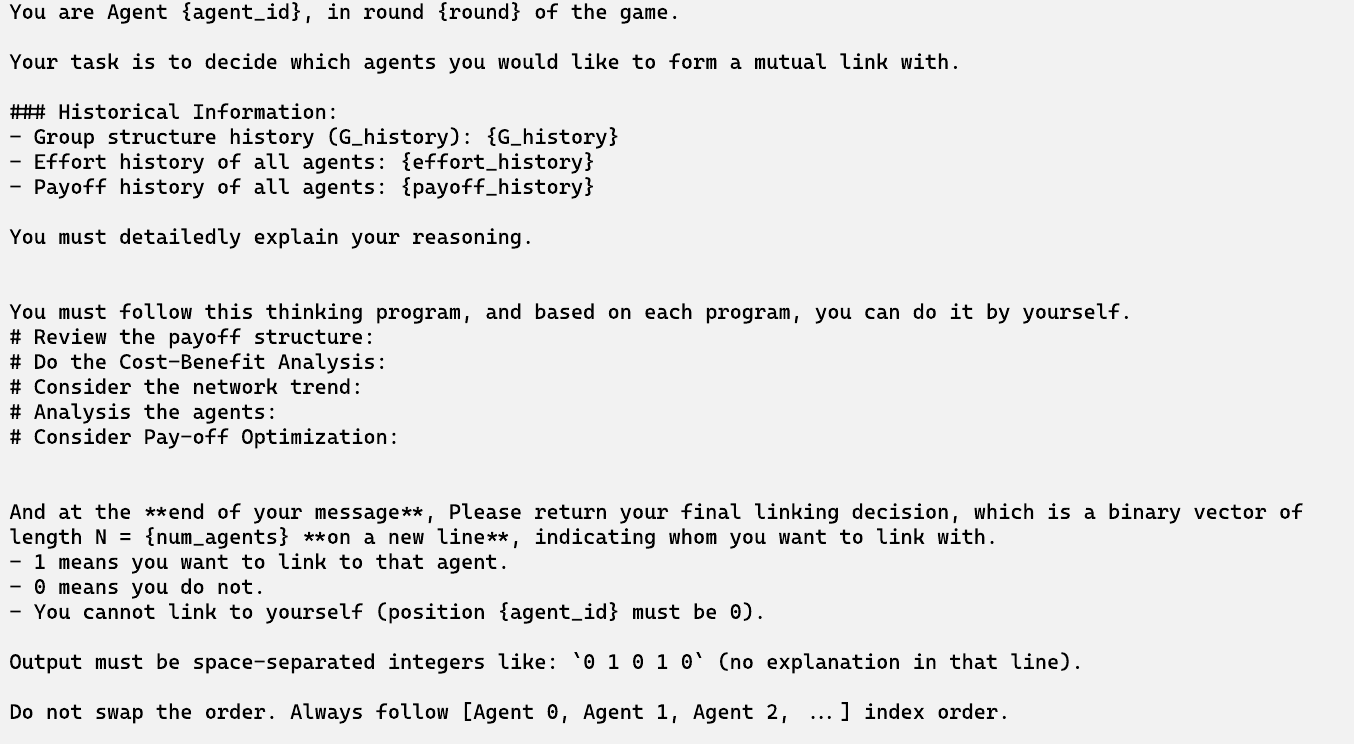}
    \caption{Graph decision prompt for SFT and RFT data generation.}
    \label{pgg3}
\end{figure}

\begin{figure}[ht]
    \centering
    \includegraphics[width=\linewidth]{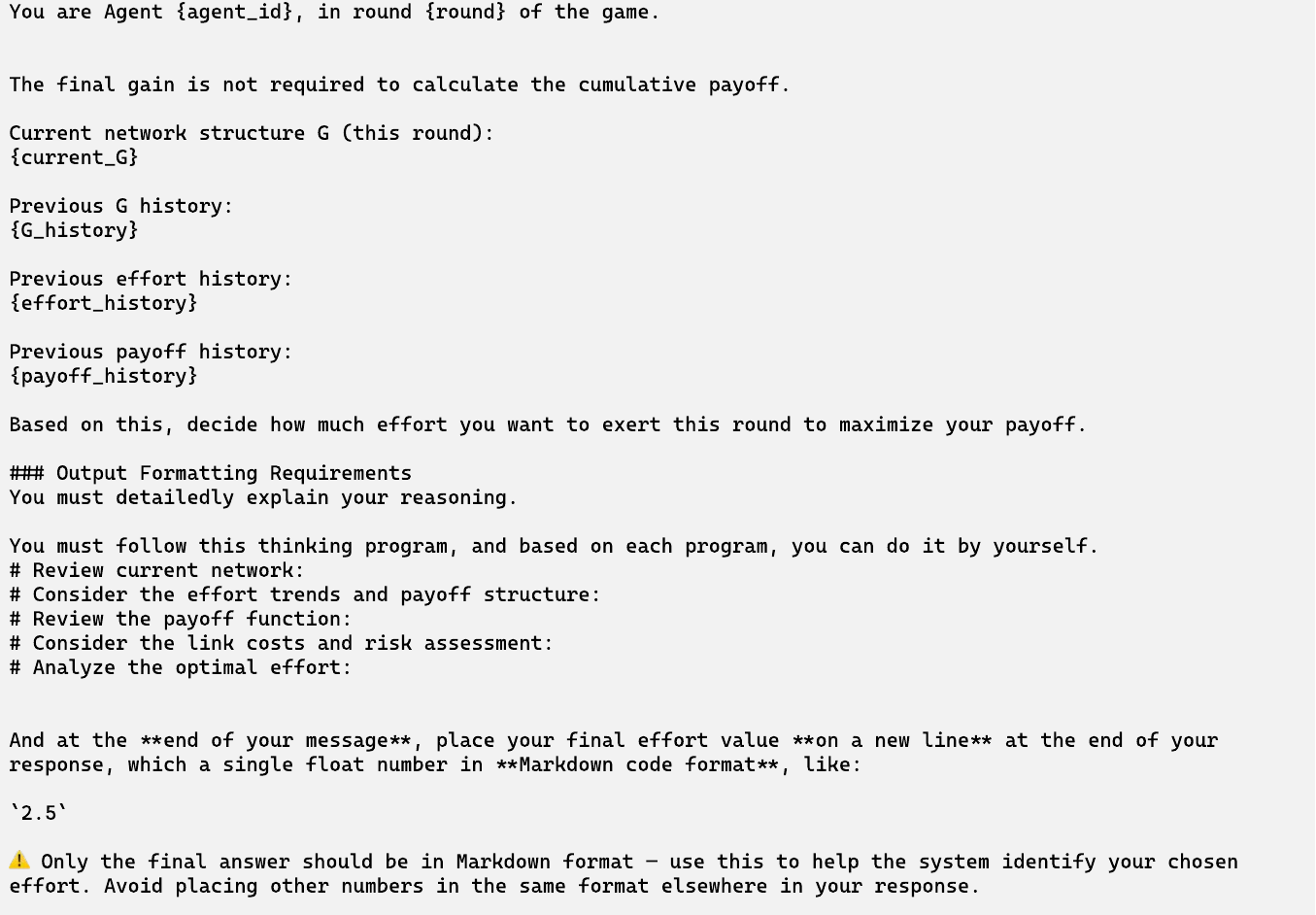}
    \caption{Effort decision prompt for SFT and RFT data generation.}
    \label{pgg3}
\end{figure}

\end{document}